\documentclass[a4paper,11pt,twocolumn,twoside]{article}
\usepackage[dvips]{graphicx}
\usepackage{sepln_en}
\usepackage{fullname}
\usepackage[utf8]{inputenc}
\usepackage{float}
\usepackage{tabularx}
\usepackage{subcaption}
\usepackage{booktabs}
\usepackage{xltabular}
\usepackage{xurl}
\usepackage{xcolor}
\def\boxit#1{%
  \smash{\color{red}\fboxrule=1pt\relax\fboxsep=2pt\relax%
  \llap{\rlap{\fbox{\vphantom{0}\makebox[#1]{}}}~}}\ignorespaces
}

\input epsf
\title{Crosslingual Argument Mining in the Medical Domain}

\author {\textbf{Anar Yeginbergen}, \textbf{Rodrigo Agerri} \\
HiTZ - Ixa, University of the Basque Country UPV/EHU\\
\{anar.yeginbergen,rodrigo.agerri\}@ehu.eus
\\
}

\seplntranstitle{Minería de Argumentos Crosslingüe en el Dominio Médico}

\seplnclave{Argumentación, Extracción de Información, Inteligencia Artificial en Medicina, Multilingualismo, Etiquetado Secuencial, Procesamiento del Lenguaje Natural.}

\seplnresumen{La tecnología basada en Inteligencia Artificial tiene una gran potencialidad para desarrollar asistentes que ayuden a profesionales médicos en la toma de decisiones, las cuales en muchos casos están basadas en el procesamiento de una gran cantidad de textos no-estructurados. En este contexto, la minería de argumentos (AM) puede ayudar a estructurar los datos textuales en componentes argumentativos y las relaciones discursivas existentes entre ellos. Sin embargo, al igual que todavía ocurre en muchas tareas de
Procesamiento del Lenguaje Natural, la gran mayoría del trabajo sobre argumentación computacional en el dominio médico se ha centrado únicamente en inglés. En este artículo investigamos varias estrategias para realizar AM en textos médicos para un idioma como el español, para el cual no
existen datos manualmente etiquetados. Nuestro trabajo muestra que traducir y proyectar automáticamente
anotaciones del inglés a un idioma de destino determinado como el español es una
forma eficaz de generar datos anotados sin necesidad de realizar anotación manual. Por otra parte, 
se demuestra experimentalmente que traducir y proyectar obtiene mejores resultados que los métodos basados en las capacidades de transferencia crosslingüe de modelos de lenguaje multilingües. Finalmente, usamos los datos automáticamente generados para español para mejorar los resultados originales en inglés, proporcionando así una estrategia de aumento de datos totalmente automática.}

\seplnkey{Argumentation, Information Extraction, AI in Healthcare, Multilinguality.}

\seplnabstract{Nowadays the medical domain is receiving much more attention in
applications involving Artificial Intelligence as clinicians decision-making is increasingly dependent on dealing with enormous amounts of unstructured textual data. In this context, Argument Mining (AM) helps to meaningfully structure textual data by identifying the argumentative components in the text and
classifying the relations between them. However, as it is the case for many
tasks in Natural Language Processing in general and in medical text processing
in particular, the large majority of the work on computational argumentation
has been focusing only on the English language. In this paper, we investigate several strategies to perform AM in medical texts for a language such as Spanish, for which no
annotated data is available. Our work shows that automatically translating
and projecting annotations (data-transfer) from English to a given target language is an
effective way to generate annotated data without costly manual intervention.
Furthermore, and contrary to conclusions from previous work for other sequence labelling tasks, 
our experiments demonstrate that data-transfer outperforms methods based on the
the crosslingual transfer capabilities of multilingual pre-trained language models (model-transfer). Finally, we show how the automatically generated data in Spanish can also be used to improve results in the original English monolingual setting, providing thus a fully automatic data augmentation strategy. Data, code, and fine-tuned models are publicly available at \url{https://huggingface.co/datasets/HiTZ/AbstRCT-ES}.
}

\firstpageno{1}

\begin{document}


\setlength\titlebox{25cm} 

\label{firstpage} \maketitle

\section{Introduction}

Clinical decision-making is an essential process in which a medical doctor, for example, 
needs to identify and diagnose a disease and prescribe a treatment
based on the patient's health condition and the results of the clinical tests.
However, it can also involve multiple challenges and add up stress, due to several
reasons. First, there may be a large diversity of symptoms, one or more of
which could be a sign of multiple diseases. Second, there exists an overwhelming
and ever-increasing amount of data from previous patients with similar
symptoms which is becoming increasingly difficult, if not impossible, to process manually. 
Lastly, the final decision should take into account the latest
results published in Evidence-based Medicine reports \cite{Sackett1996EvidenceBM}. 

These challenges highlight the importance of medical professionals to be supported by AI-based technology able to extract, from the huge quantity of unstructured textual data available for the different diseases and treatments, relevant and timely information structured in a manner that is suitable to be easily analyzed.

Argument mining (AM) is a field of Natural Language Processing (NLP) that focuses on
extracting argumentative structures from unstructured textual content. While there are a large number of theories on argumentation, in this work we
follow the model proposed by \namecite{stab2014annotating} according to which the main objective of AM is to generate hierarchical argumentative tree structures by automatically identifying in running text the types of argument components (e.g., \emph{evidences} or \emph{claims}) and their boundaries (a sequence labelling task) and the relations held between them (e.g., \emph{support} or \emph{attack}), addressed as pair-wise classification. Given that such argumentative structures constitute the basis of evidence-based reasoning applications, research on AM for the medical domain has the potential of facilitating the development of enabling technology to help in clinical decision-making.

A number of works have been proposed to automatically detect,
classify and assess the quality of argumentative structures for various specific domains such as law
\cite{mochales2009creating}, biomedicine \cite{mayer2021enhancing,accuosto2021argumentation}, 
reviews \cite{li2017crowdsourcing} and persuasive essays \cite{stab2014annotating,stab-gurevych-2017-parsing} but the large majority of them have been focused on English, especially for the medical domain.

Although some researchers have translated Persuasive Essays to study crosslinguality in AM for languages such as Chinese, German, French, Spanish, or Portuguese \cite{eger-etal-2018-cross,Sousa2021CrossLingualAP}, the generated data and method are not suitable for our purposes. First, translation and label projection for argument components have been done via manual translation or by applying fully automatic methods. However, while interesting, manually translating and projecting the labels is a route to avoid in the \emph{data-transfer} approach since the objective is to reduce manual work to a minimum by applying Machine Translation (MT). Second, the automatic projection of labels should be manually checked \emph{at least} for evaluation \cite{garcia2022model} using both \emph{model-transfer} and \emph{data-transfer} methods. Finally, for the medical domain the only existing AM dataset is in English, which means that for a major language such as Spanish no annotated dataset for AM in the medical domain is currently available. In order to address these issues, this work empirically investigates optimal crosslingual transfer techniques to perform AM in medical texts for a target language such as Spanish for which no manually annotated data is available.


In this work we leverage AbstRCT, at the time of writing the only existing labeled dataset in English for AM in the medical domain \cite{mayer2021enhancing}, to perform crosslingual transfer about argument
components and relations from English to Spanish. More specifically, we experiment with: (i) \emph{model-transfer} in which a multilingual Masked Language Model (MLM) is fined-tuned in English and the predictions are generated in Spanish; (ii) \emph{data-transfer} techniques based on machine translating and projecting the argument component labels to generate annotated data for Spanish with a minimum effort and, (iii) \emph{automatic data augmentation} by exploiting the multilingual capabilities of MLMs with the aim of improving results in the original source English gold-standard data. Summarizing, the main contributions of this paper are the following:\\
\noindent - We present the first Spanish dataset annotated with the argument
		components and relations at the sequence level for the medical domain.\\
\noindent - We perform the first experiments comparing \emph{model-transfer} and
		\emph{data-transfer} methods for AM in order to establish which
		strategy works best when no annotated data is available for a target
		language. Contrary to previous work for other sequence labelling tasks \cite{garcia2022model}, our results demonstrate the superiority of \emph{data-transfer} for argument component detection in the medical domain.\\
\noindent - We show how \emph{data-transfer} can easily and
		effectively be used to perform fully automatic data augmentation to
		improve results in the original English evaluation dataset.\\
\noindent - We establish that training in \emph{data-transfer} with projected labels (with no manual correction) obtains similar performance to training with manually-checked labels, outperforming also \emph{model-transfer} results. This is a highly encouraging outcome of our work, as it means that we should be able to discard any manual correction of the projected labels, making the whole \emph{data-transfer} process fully automatic.\\
\noindent - The generated Spanish annotated data (both with automatically projected labels and its manually revised version) are publicly available to encourage crosslingual research in argument mining and to facilitate the reproducibility of results\footnote{\url{https://huggingface.co/datasets/HiTZ/AbstRCT-ES}}.


\section{Related Work}\label{sec:related-work}

One of the pioneer works with a clear influence in computational argumentation proposed 
a number of functional roles for arguments based on the manner in which the conclusion is made based on textual evidence: evidence, warrant, backing, qualifier, rebuttal, and claim \cite{toulmin1958uses}. Furthermore, another important contribution argued that the
most important relationship types between argument components are those of \emph{support}
and \emph{attack} \cite{peldszus2013argument}. Interestingly, they also defined five types of
argumentation graphs according to the number and type of existing relations between the argument components: one claim having relations with multiple premises, a claim followed by another claim, etc.

Aiming to develop a fully computational AM approach, Stab and Gurevych \shortcite{stab2014annotating,stab-gurevych-2017-parsing} proposed a model in which the objective of AM was to generate hierarchical structures to represent the argument components (\emph{claims} and \emph{premises}) in a manually annotated dataset of Persuasive Essays and their relations (\emph{support} and \emph{attack}). Their model has been widely adopted in NLP, also for the medical domain \cite{stab-gurevych-2017-parsing,eger-etal-2018-cross,accuosto2021argumentation,Sousa2021CrossLingualAP,mayer2021enhancing}.

\subsection{AM in the Medical Domain} \label{arg_mininig_medical_domain}

Argument mining can be very valuable in the medical domain, particularly in
Evidence-based Medicine where argumentative explanations are given in order to
justify or discard a given diagnosis or treatment \cite{mayer2021enhancing}. However, the few previous
works on argument mining in the medical domain are all focused on English. In fact, to the best of our knowledge, there is only one dataset, AbstRCT, which provides annotated argument components and relations in the medical domain \cite{mayer2021enhancing}. This is partially
due to the inherent difficulty of obtaining medical data to start with, but
also because of the cost and complexity of manually annotating argumentative
structures. \namecite{mayer2021enhancing} present a large battery of experiments using various English MLMs specific to the medical domain \cite{beltagy2019scibert,lee2019biobert}. They also demonstrate that combining Transformer-based models with CRF, LSTM or GRU helped to improve results.

With respect to other proposals, \namecite{green2014argumentation} provided an analysis of arguments
in biomedical data and defined argumentation schemes and inter-argument
relationships. \namecite{alamri2016corpus} created a corpus using research
abstracts of studies considered in systematic reviews related to cardiovascular
diseases where the objective was solely to identify contradictory claims. \namecite{shankar2006medical} developed a tool for healthcare where one of the functionalities is extracting evidence for any treatment-related claims based on an argumentative structure defined by \namecite{toulmin1958uses}. Finally, \namecite{accuosto2021argumentation} generated a dataset of arguments annotated at sentence level for bioscientific documents.

\subsection{Crosslingual Transfer} \label{cross_lingual_seq}


\begin{table*}
\begin{center}
\begin{tabular}{l|c|c|c|c|c} 
 \toprule
Split & \#prem & \#claim & \#support & \#attack & \#no-rel \\ 
 \midrule
 Train - Neoplasm & 1535 & 730 & 1194 & 200 & 12892\\ 
 Dev - Neoplasm & 218 & 108 & 185 & 30 & 1815 \\
 Test - Neoplasm & 438 & 248 & 359 & 60 & 3961 \\
 Test - Glaucoma & 404 & 190 & 317 & 29 & 2986 \\
 Test- Mixed & 388 & 212 & 296 & 24 & 3012 \\
\midrule
Total & 2983 & 1488 & 2351 & 343 & 24666 \\
 \bottomrule
\end{tabular}
\caption{Distribution of argument components and relations in the AbstRCT dataset.}
\label{table:abstrct}
\end{center}
\end{table*}

Crosslingual transfer techniques for sequence labelling aim at transferring knowledge from rich source languages (usually English) to other less resourced languages with the aim of mitigating the lack of annotated data in the target language \cite{david2001inducing}. 

Most of the work was initially based on leveraging some kind of parallel data or representations like crosslingual embeddings for tasks such as POS tagging \cite{david2001inducing,gaddy2016ten},
named entity recognition (NER) \cite{gaddy2016ten,yang2017transfer,agerri2018building} and extractive question answering \cite{lewis-etal-2020-mlqa,artetxe-etal-2020-cross}, among others \cite{chen-etal-2019-multi-source,Liu2020OnTI}.


\namecite{garcia2022model} provide an exhaustive comparison between \emph{model-transfer} and \emph{data-transfer} approaches on a variety of sequence labelling tasks, datasets, and languages.
They conclude that \emph{model-transfer} using large MLMs such as
XLM-RoBERTa-large \cite{Conneau2020UnsupervisedCR} outperforms \emph{data-transfer}
(translate and project) methods in every evaluation setting.

For AM there have also been works applying translation and label projection from English to other target languages. In this case, the objective was to compare the downstream task results when training with manually translated and projected data with respect to fully automatic one \cite{eger-etal-2018-cross,Sousa2021CrossLingualAP}.

In order to study crosslingual transfer for AM in medical texts, in this paper we develop the first Spanish corpus annotated with argument components and relations for the medical domain. This allows us to present the first experiments on \emph{model-transfer} and \emph{data-transfer} methods for AM. 
Surprisingly, and contrary to previous conclusions for other sequence labelling tasks \cite{garcia2022model}, our results demonstrate the superiority of \emph{data-transfer} for AM in the medical domain.

\section{Translate and Project}\label{sec:data-transfer}

\begin{figure*}
    \centering
    \includegraphics[width=1.0\textwidth]{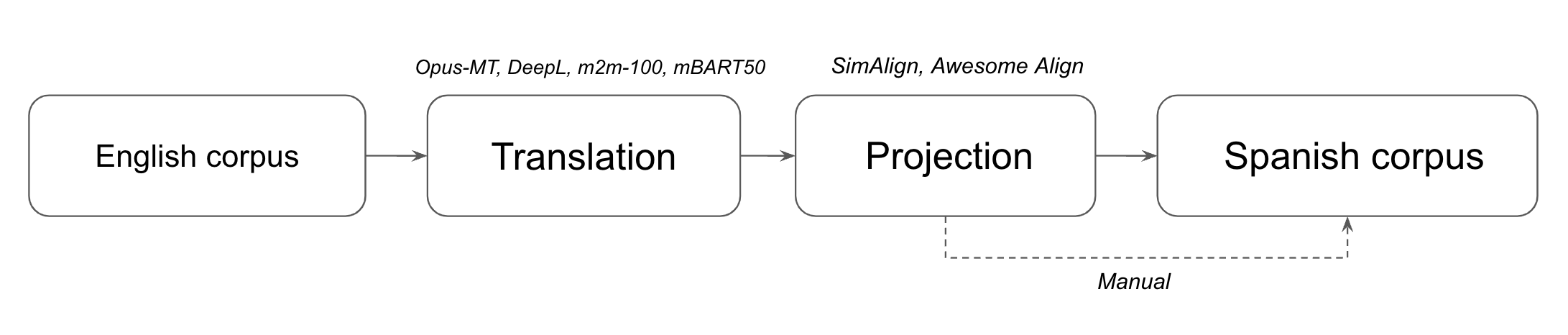}
    \caption{The process of creating Spanish data from English.}
    \label{fig:en_to_es}
\end{figure*}

In this section we describe the steps undertaken in order to generate a Spanish version of AbstRCT \cite{mayer2021enhancing} by applying MT and label projection for argument components (\emph{claims} and \emph{premises}). The process is shown in Figure \ref{fig:en_to_es}. First, AbstRCT is translated and the argument component labels are automatically projected by exploiting word alignments generated by SimAlign  \cite{sabet2020simalign} and \textsc{awesome} \cite{dou2021word}. Finally, the projected labels were manually checked to ensure trustworthy evaluation results.

\subsection{AbstRCT Dataset}

The dataset of Randomized Clinical Trials (RCT) consists of randomized controlled trials retrieved from the MEDLINE database via PubMed search\footnote{https://pubmed.ncbi.nlm.nih.gov/7850561/}. The trials are annotated with argument components and argumentative relations. The corpus contains paragraphs on five types of diseases: neoplasm, glaucoma, diabetes, hepatitis B, and hypertension. Overall there are 500 neoplasm, 100 glaucoma, and 100 mixed (20 from each of the 5 diseases) abstracts. The corpus includes the annotations for the argument components and relations in separate files. Table \ref{table:abstrct} provides the distribution of argument components and relations. Their definition follows previous work \cite{stab-gurevych-2017-parsing} but adapted to the medical domain.

\textbf{Claims} are concluding statements about the outcome of the study. In the medical domain it may be a diagnosis or a treatment.

\textbf{Premises} correspond to an observation or measurement in the study (ground truth), which supports or attacks another argument component, usually a claim. It is important that they are observed facts, therefore, credible without further evidence.

One document may contain multiple claims and premises that \emph{support} or \emph{attack} each other. Example (1) below presents an example of \emph{claims} marked in bold with subscript $C_n$, and of \emph{premises} in italics with subscript $P_n$.

\textbf{Attack relations} occur when the source component contradicts the
target or when it states that some observation had no statistical significance.
In Example (1), Premise$_2$
(\textit{P$_2$}) (partially) attacks Claim$_2$ (\textit{C$_2$}), namely, it
qualifies the conditions of the study without fully objecting to it. 

\textbf{Support relations} provide a justification from the source to the
target. Example (1) contains several support links
between arguments, for instance, Premise$_1$ (\textit{P$_1$}) supports
Claim$_1$ (\textit{C$_1$}) with numerical evidence to verify the statement. 

\noindent (1)
``Results of the pretest showed that many patients lacked knowledge about pain and pain
management. The majority of pain topics had to be discussed. \textbf{[The Pain
Education Program proved to be feasible]}$_{\textbf{$C_1$}}$: \emph{[75.0\%
of the patients had read the entire pain brochure, 55.7\% had listened to the
audio cassette, and 85.6\% of pain scores were completed in the pain
diary]}$_{\textbf{$P_1$}}$. \textbf{[Results showed a significant increase in
pain knowledge in patients who received the Pain Education Program and a
significant decrease in pain intensity]}$_{\textbf{$C_2$}}$. \emph{[However,
pain relief was mainly found in the intervention group patients without
district nursing]}$_{\textbf{$P_2$}}$. \textbf{[It can be concluded that the
tailored Pain Education Program is effective for cancer patients in chronic
pain]}$_{\textbf{$C_3$}}$.''

\subsection{Machine Translation}

Four MT systems were tested to translate the AbstRCT dataset into Spanish. Before proceeding to translate the full dataset we performed a pilot by translating a small set of sentences using m2m-100 \cite{fan2021beyond}, mBART \cite{tang2020multilingual}, OPUS-MT \cite{tiedemann2020opus} and DeepL\footnote{https://www.deepl.com/}. In order to obtain a ranking of the MT systems, the output of each model was then blindly evaluated by two native Spanish-speaking experts in medical NLP. DeepL was, overall, the best-performing system, and OPUS-MT was the second-best-performing. The kappa inter-annotator agreement (IIA) in choosing the best MT system was around $\sim$70\%.

\begin{figure*}
    \centering
    \includegraphics[width=1.0\textwidth]{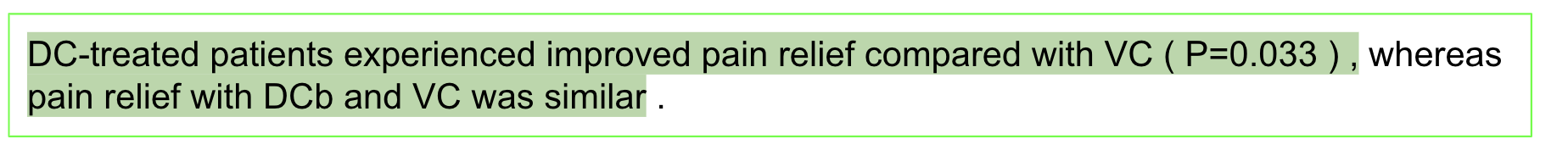}
    \includegraphics[width=1.0\textwidth]{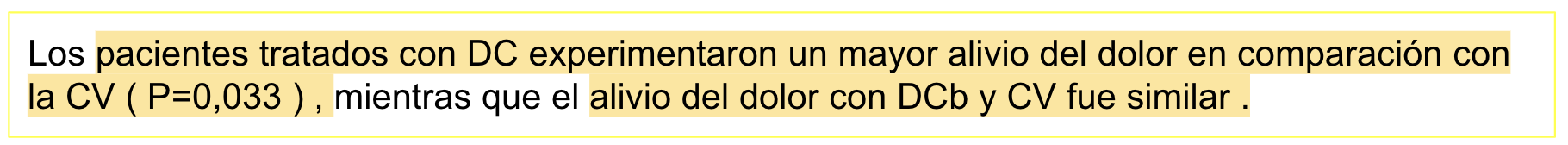}
     \includegraphics[width=1.0\textwidth]{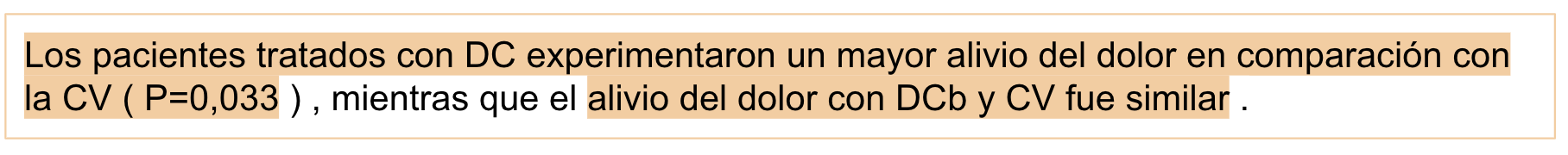}
    \caption{Alignment issues to project two \emph{premise} argument component labels from English (top) to Spanish with \textsc{awesome} (middle) and SimAlign (bottom).}
    \label{fig:alignments}
\end{figure*}

A noticeable issue in translating scientific and medical data was the difficulty in dealing with abbreviations. Thus, the ability of MT systems to deal with them was one of the criteria in the selection of the most accurate model, besides linguistic coherence and correctness. Once the MT system (DeepL) was chosen, the next step consisted of projecting the argument component labels (\emph{claims} and \emph{premises}) from the original English annotated data to the automatically translated Spanish data. Note that label projection is only challenging for argument components as they consist of spans of texts that need to be aligned from the source to the target language (as illustrated in Figure \ref{fig:alignments}), while \emph{support} and \emph{attack} relation classification is modeled as pairwise classification in which there is a 1-1 relationship which means that there is no need to perform label projection.

\subsection{Label Projection}

Following previous work on crosslingual transfer \cite{david2001inducing,agerri2018building} we project the argument component labels via word alignments. In this work we use state-of-the-art methods for word alignment based on Transformer-based multilingual contextualized embeddings, namely, \emph{SimAlign} \cite{sabet2020simalign} and \textsc{awesome} \cite{dou2021word}. Once you obtain the word alignments then we need to project the sequences of full argument components, an extra step required which is not covered by the word alignment algorithms. Instead of coming up with a new method, we apply the automatic annotation projection algorithm, Easy Label Projection, developed by \namecite{garcia2022model}. However, as Easy Label Projection was developed for tasks such as NER or Opinion Target Extraction which are characterized by shorter and homogeneous sequences, we had to implement several changes to adapt it to the projection of long and heterogeneous argumentative sequences such as \emph{claims} and \emph{premises}.

While SimAlign allows projecting labels without any parallel data by extracting alignments from similarity matrices of multilingual embeddings, \textsc{awesome} requires parallel data, so we decided to use the English-Spanish parallel biomedical corpus.\footnote{\url{https://github.com/biomedical-translation-corpora/corpora}}

We generated three versions of AbstRCT in Spanish according to the projection method: (i) \emph{automatic} projections, (ii) \emph{post-processed} projections which programmatically corrected the \emph{automatic} projections and, (iii) \emph{manually} corrected projections.

\subsubsection{Automatic projections}

\begin{table*}
\small{
\begin{center}
\begin{tabular}{l|c|c|c|c} 
 \cmidrule{2-5}
 & train\_awesome & train\_simalign & dev\_awesome & dev\_simalign \\
\midrule
\# total & 4405 & 4405 & 680 & 680 \\
\# full O's  & 2345 & 2345 & 377 & 377 \\
\# full component & 1752 & 703 & 257 & 257 \\
\# post-processed & 800 & 88 & 95 & 11 \\
\# manual-corrections & 140 & 194 & 20 & 25 \\
\bottomrule
\end{tabular}
\caption{Train and development Neoplasm data in AbstRCT translated with DeepL. From top to down: Total number of argument components projected; number of sentences composed entirely of O tokens; number of sentences forming an argument component; number of post-processed sequences and number of manually corrected sequences.}
\label{table:postprocessed_projection_train_dev}
\end{center}
}
\end{table*}

\begin{table*}
\small{
\begin{center}
\begin{tabular}{l|c|c|c|c|c|c}
 \cmidrule{2-7}
 &  neoplasm\_a & neoplasm\_s & glaucoma\_a & glaucoma\_s & mixed\_a & mixed\_s \\
\midrule
\# total & 1252 & 1252 & 1248 & 1248 & 1147 & 1147 \\
\# full O's  & 630 & 630 & 692 & 682 & 591 & 591 \\
\# full component & 518 & 518 & 498 & 506 & 476 & 480 \\
\# post-processed  & 242 & 92 & 167 & 51 & 203 & 90 \\
\# manual-corrections & 51 & 26 & 26 & 14 & 47 & 26 \\
\bottomrule
\end{tabular}
\caption{Statistics of projected components for the DeepL translated test sets in AbstRCT; \_a refers to \textsc{awesome} while \_s corresponds to SimAlign.}
\label{table:postprocessed_projection}
\end{center}
}
\end{table*}

The middle alignment in Figure \ref{fig:alignments} produced by applying \textsc{awesome} and Easy Label Projection \cite{garcia2022model} shows the the difficulty for this method to correctly bracket the beginning of the \emph{premise} sequences by including the two articles in the argument component. Another typical issue was erroneously including the final punctuation in the sentence as part of the argument component.

While SimAlign handled some articles at the beginning of argument component relatively better, it is difficult to find patterns in the misaligned cases. Thus, by looking at the last example in Figure \ref{fig:alignments}, it can be seen that SimAlign correctly projected the first article, but not the `el' article just after `mientras que'. Furthermore, the closing bracket and coma were also incorrectly projected although the final full stop was correctly left outside of the argument component.\footnote{It is worth mentioning that in the original English AbstRCT punctuation is not consistently annotated.}

\subsubsection{Post-processed projections}

After identifying frequent mistakes according to some patterns, we were able to implement a simple method to automatically correct them. Furthermore, and taking into account that many sequences were full sentences, we applied a post-processing step to directly project the full sentence into Spanish, without requiring the use of word alignments and Easy Label Projection. This allowed us to automatically correct a large number of incorrectly projected sequences, as shown by \#post-processed results in Tables  \ref{table:postprocessed_projection_train_dev} and \ref{table:postprocessed_projection}.

\subsubsection{Manual corrections}

After the post-processing step, we manually checked the rest of the projected sequences to correct any possible misalignments. During this manual revision punctuation at the end of an argument component was systematically included as part of the sequence, even if this was not the case in the original English data.

The final number of manually corrected sequences is reported in the last row \#manual-correction of Tables \ref{table:postprocessed_projection_train_dev} and \ref{table:postprocessed_projection}. As it can be seen, after post-processing the number of corrections are quite low, reducing the manual work to generate a Spanish dataset for AM in the medical domain to a minimum.

\begin{table*}
\footnotesize{
\begin{center}
\begin{tabular}{c|ccc|ccc|ccc} 
 \toprule
  Model & & Neoplasm &&& Glaucoma &&& Mixed & \\
  \midrule
  & F1 & F1-C & F1-P & F1& F1-C & F1-P & F1 & F1-C & F1-P \\ 
 \midrule
SciBERT+GRU+CRF  & 82.41 & 75.84 & 91.11 & 83.97 & 82.89 & 91.68 & 82.40 & 78.21 & 91.35 \\ 
BioBERT+GRU+CRF  & 80.85 & 73.99 & 90.59 & 83.95 & 83.52 & 91.72 & 82.41 & 78.26 & 92.02 \\ 
BERT+GRU+CRF     & 82.68 & 76.23 & 89.90 & 82.22 & 79.07 & 89.07 & 82.68 & 77.98 & 89.61 \\ 
\boxit{15.2cm}mBERT+GRU+CRF    & 82.36 & 74.89 & 89.07 & 80.52 & 75.22 & 84.86 & 81.69 & 75.06 & 88.57 \\ 
SciBert+LSTM+C RF & 81.99 & 75.58 & 91.23 & 83.06 & 81.87 & 91.76 & 81.93 & 77.23 & 91.52 \\
\bottomrule
\end{tabular}
\caption{Argument component detection of the source AbstRCT English data.}
\label{table:arg_comp_res}
\end{center}
}
\end{table*}

\section{Experimental Setup}\label{sec:experimental-setup}

During the label projection described in the previous section we generated six versions of the Spanish AbstRCT dataset for argument component detection: three different projection types (automatic, post-processed and manual) with two different aligners (SimAlign and \textsc{awesome}). Furthermore, as no word alignments are required, we also directly generated one version for relation classification. In order to establish the optimal strategy to perform AM, namely, \emph{model-transfer} or \emph{data-transfer}, when no annotated data is available for a target language, we run the following experiments: 

\begin{itemize}
    \item \emph{model-transfer}: fine-tune multilingual MLMs in English and test in Spanish.
    \item \emph{data-transfer}: fine-tune and test the MLMs using the AbstRCT Spanish dataset generated by MT and label projection. Evaluation is always performed on manually corrected data.
    \item \emph{multilingual}: simple data augmentation using multilingual MLMs by combining the English and Spanish training data and evaluate in each of the target languages.
\end{itemize}

\subsection{Models}

For the experiments we use the architecture proposed in the original AbstRCT experiments \cite{mayer2021enhancing} although instead of using monolingual English MLMs as a backbone, for crosslingual transfer we will need to deploy a multilingual MLM. \namecite{mayer2021enhancing} experimented with adding sequence labelling components such as Conditional Random Fields (CRF), Recurrent Neural Network (RNN), or Gate Recurrent Units as a last layer on top of a Transformer model \cite{vaswani2017attention}. The intuition is that since the length of argument components is considerably long, adding a sequence labelling layer should help to better capture long dependencies between the tokens included in an argument component.

Results obtained showed that adding both a GRU and a CRF component to English MLMs specific to the medical domain such as BioBERT \cite{lee2019biobert} and SciBERT \cite{beltagy2019scibert} were the optimal options for argument component detection.

After the release of encoder-only MLMs for English caused a
paradigm-shift in NLP research, multilingual MLMs models such as multilingual BERT and XLM-RoBERTa, trained to work on 100 languages, were published, obtaining excellent results on both monolingual and, especially, on multilingual and crosslingual settings \cite{Pires2019HowMI,Wu2020AreAL,Conneau2020UnsupervisedCR}. Therefore, as our objective is to study crosslingual transfer for AM, we decided to use a multilingual MLM as a backbone. Quite surprisingly, and contrary to previous results on Spanish and crosslingual transfer benchmarks for encoder-only MLMs \cite{Agerri2022LessonsLF,garcia2022model,Conneau2020UnsupervisedCR}, preliminary experimentation on argument component detection established that multilingual BERT (mBERT) \cite{Devlin2019BERTPO} obtained better results than XLM-RoBERTa \cite{Conneau2020UnsupervisedCR}. Multilingual BERT was trained with a batch size of 256 and 512 sequence length for 1M steps, using both the MLM and Next Sentence Prediction (NSP) tasks.

After hyperparameter tuning on the English development data, the best results were obtained with 3 epochs, 5e-5 of learning rate, 32 batch size and 256 of maximum sequence length.

\section{Results}\label{sec:results}

Before proceeding to run the experiments on \emph{model-} and \emph{data-transfer}, we first tested mBERT as a backbone on the AM architecture originally designed for the English AbstRCT experiments \cite{mayer2021enhancing}, as described above. We also fine-tuned the best English MLMs models following the method applied for English \cite{mayer2021enhancing}. We report the results in Table \ref{table:arg_comp_res} (F1 score is an average of F1-Claim (F1-C) and F1-Premise (F1-P) calculated at token level, as evaluated by the system used for the English AbstRCT).

F1-scores for Neoplasm show that mBERT obtains competitive results with respect to specialized monolingual models such as SciBERT and BioBERT. However, mBERT results are slightly worse when evaluated out-of-domain in Glaucoma and Mixed test data. In any case, this baseline shows that mBERT performs competitively on the English original data which makes it a good candidate to perform crosslingual and multilingual experiments.

\begin{table}[h]
\footnotesize{
\begin{center}
\begin{tabular}{c|c|c|c} 
 \toprule
  Model & Neoplasm & Glaucoma & Mixed \\
  \midrule
BERT    & 66.97 & 57.04 & 69.32 \\ 
SciBERT & 70.31 & 65.75 & 71.31 \\
BioBERT & 55.84 & 59.23 & 56.17 \\ 
\boxit{6.2cm}mBERT   & 65.71 & 59.92 & 67.88 \\
 \bottomrule
\end{tabular}
\caption{F-1 score for relation classification on the source AbstRCT English data.}
\label{table:rel_clf_baseline}
\end{center}
}
\end{table}

We perform the same baseline experiments for relation classification and report the results in Table \ref{table:rel_clf_baseline}. While the differences are a bit larger than for argument component detection in Table \ref{table:arg_comp_res}, especially with respect to SciBERT, mBERT remains competitive when compared to most monolingual English models.

\begin{table*}
\footnotesize{
\begin{center}
\begin{tabular}{c|ccc|ccc|ccc} 
 \toprule
  Test & & Neoplasm &&& Glaucoma &&& Mixed & \\
  \midrule
  & F1 & F1-C & F1-P & F1 & F1-C & F1-P & F1 & F1-C & F1-P \\ 
  \midrule
  \multicolumn{10}{c}{\textbf{model-transfer}}\\
  \midrule
    DeepL + SimAlign (ES) & 80.50 & 71.56 & 86.73 & 77.60 & 72.33 & 81.60 & 79.62 & 68.99 & 85.96 \\ 
    DeepL + Awesome (ES) & 80.34 & 71.54 & 86.77 & 77.51 & 72.30 & 81.59 & 79.57 & 69.36 & 85.92 \\
 \midrule
  \multicolumn{10}{c}{\textbf{data-transfer}} \\
\midrule
   DeepL + SimAlign (ES) & 83.57 & 75.95 & 90.01 & 80.83 & 75.44 & 86.11 & 82.62 & 74.62 & 88.81 \\ 
   DeepL + Awesome (ES) & 83.40 & 77.11 & 89.18 & 81.11 & 76.16 & 87.35 & 81.88 & 73.71 & 88.50 \\ 
 \midrule
 \multicolumn{10}{c}{\textbf{multilingual}} \\
 \midrule
 English  & \boxit{0.7cm}83.51 & 73.42 & 89.38 & \boxit{0.7cm}85.31 & 81.05 & 86.73 & \boxit{0.7cm}83.63 & 74.98 & 89.25 \\
    DeepL + SimAlign (ES) & \boxit{0.7cm}84.35 & 76.64 & 88.43 & \boxit{0.7cm}84.54 & 78.67 & 87.24 & \boxit{0.7cm}83.90 & 73.46 & 88.87\\
                                 
    DeepL + Awesome (ES) & \boxit{0.7cm}84.58 & 76.77 & 88.61 & \boxit{0.7cm}84.62 & 78.71 & 87.23 & \boxit{0.7cm}84.03 & 73.99 & 88.92 \\ 
    \bottomrule
\end{tabular}
\caption{Results for argument component detection of \emph{model-transfer}, \emph{data-transfer} and \emph{multilingual} experiments. Spanish data corresponds to manually corrected projections; in red best overall results per test set.} 
\label{tab:crosslingual_results}
\end{center}
}
\end{table*}

\paragraph{Crosslingual Transfer for Argument Components}


Table \ref{tab:crosslingual_results} compares the performance in the different evaluation settings. The first noticeable result is that for argument component detection, \emph{data-transfer} (training on data obtained by MT and label projection from English to Spanish) clearly outperforms \emph{model-transfer} (training on English and predicting in Spanish). This result differs from previous works on crosslingual transfer for sequence labelling in which \emph{model-transfer} was clearly superior \cite{garcia2022model}. We hypothesized that MT in the medical domain may introduce less translation artifacts than for other more open-ended domains.

Secondly, simple data augmentation by training on the English data together with their translated and projected counterpart in Spanish results in important performance gains. In fact, it outperforms the original English results reported in Table \ref{table:arg_comp_res} (also against the monolingual domain-specific models) and it also helps to improve results over the monolingual Spanish \emph{data-transfer} setting. 
It should also be noted that the improvements are particularly large when mBERT is evaluated out-of-domain, namely, on the Glaucoma and Mixed test sets.

Another aspect worth mentioning is that we obtain roughly the same results with both types of word alignments, which means that the process of \emph{post-processing} and \emph{manual correction} was robust enough to guarantee the same results regardless of the aligner used.

Summarizing, the best results for both languages are obtained with the multilingual data augmentation strategy, also surpassing the models trained with gold-standard English data. Therefore, \emph{data-transfer} proves to be the most effective strategy to perform argument component detection in the medical domain whenever no annotated data for the target language is available.

\paragraph{Relation classification}

All the experiments that were done for argument component detection were also applied to the classification of argument relations. Results are reported in Table \ref{table:rel_clf_es}.

\begin{table*}
\small{
\begin{center}
\begin{tabular}{l|c|c|c} 
 \toprule
  Model & Neoplasm & Glaucoma & Mixed \\
  \midrule
train: EN+ES $\to$ test: EN & 65.55 & 58.79 & 67.82  \\ 
train: EN+ES $\to$ test: ES & 62.55 & 58.60 & 65.74  \\
train: EN $\to$ test: ES & 62.45 & 55.92 & 65.02  \\
train and test: ES & 63.25 & 54.35 & 65.40   \\ 
 \bottomrule
\end{tabular}

\caption{F-1 scores for relation classification in crosslingual settings.}
\label{table:rel_clf_es}
\end{center}
}
\end{table*}
\begin{table*}
\begin{center}
\resizebox{16cm}{!}{
\begin{tabular}{c|ccc|ccc|ccc} 
 \toprule
  Model & & Neoplasm &&& Glaucoma &&& Mixed & \\
  \midrule
  & auto & post & manual & auto & post & manual & auto & post & manual \\ 
 \midrule
    DeepL + SimAlign & 75.87 & 79.99 & 80.50 & 75.64 & 77.30 & 77.60 & 75.25 & 79.25 & 79.62 \\ 
    DeepL + Awesome  & 70.07 & 79.12 & 80.34 & 71.39 & 76.77 & 77.51 & 69.85  & 78.21 & 79.57 \\
 \bottomrule
\end{tabular}
}
\caption{F1-macro for argument component detection in \emph{model-transfer} setting; training in English and evaluating with projections (i) automatic; (ii) post-processed and (iii) manually corrected.}
\label{table:compare_projections}
\end{center}
\end{table*}

Here, similar to the argument components, the models trained on multilingual data also obtain the best overall results across the three test sets and \emph{data-transfer} outperforms \emph{model-transfer}, especially on the Glaucoma set. Furthermore, although very close, in this case, the multilingual results do not improve over the results in English obtained with gold standard data (reported in Table \ref{table:rel_clf_baseline}).

An important point to mention is that, as shown in Table \ref{table:abstrct}, the relation classification corpus is extremely imbalanced, which may be why classification results are substantially lower. Another reason is the lack of context in the task itself, given that in many cases it is extremely difficult to distinguish relations given only two sentences, without further context. Thus, we believe that relation classification in the medical domain cannot be straightforwardly determined based only on local textual information. Instead, it may require more complex structures and additional insights from the data.

\section{Discussion}\label{sec:discussion}

As an additional analysis, we compare the results when evaluating \emph{model-transfer} on the three types of projected data generated: (i) automatically projected labels, (ii) post-processed projection and (iii) manually checked labels. Table \ref{table:compare_projections} reports the results. The most interesting outcome is that, while evaluating on automatic projections substantially degrades the final performance, the results obtained by evaluating on the post-processed data are practically the same as when evaluating on manually checked projections. This means that \emph{data-transfer} can be performed fully automatically while obtaining reliable evaluation results, which is an extra argument in favour of this method to perform AM in the medical domain.

Overall, we can see how the results obtained by mBERT in this setting are high enough and they get better with each improvement introduced in the evaluation sets. The main issue with the automatic projections was that the alignment boundaries were different to those in the English gold training data (many errors were committed by not including a Spanish article as part of the argument), therefore the prediction scores, as expected, are noisy and unreliable.

\begin{table*}[h]
\begin{center}
\resizebox{15cm}{!}{
\begin{tabular}{c|ccc|ccc|ccc} 
 \toprule
    & & Neoplasm &&& Glaucoma &&& Mixed & \\
  \midrule
Test  & F1 & F1-C & F1-P & F1 & F1-C & F1-P & F1 & F1-C & F1-P \\ 
\midrule
\multicolumn{10}{c}{\textit{Train: DeepL + SimAlign - Automatic Projections}} \\ 
\midrule
    DeepL + SimAlign    & 82.33 & 75.90 & 87.64 & 80.94 & 75.36 & 85.97 & 81.81 & 74.99 & 88.07 \\ 
    DeepL + Awesome     & 81.98 & 75.27 & 86.99 & 80.84 & 75.31 & 85.96 & 81.73 & 75.36 & 88.03 \\    
 \midrule
 \multicolumn{10}{c}{\textit{Train: DeepL + Awesome - Automatic Projections}} \\ 
\midrule
    DeepL + SimAlign     & 72.78 & 72.66 & 87.03 & 74.21 & 72.82 & 84.74 & 72.40 & 70.17 & 87.99 \\ 
    DeepL + Awesome      & 72.50 & 72.62 & 87.00 & 74.20 & 72.77 & 84.75 & 72.43 & 70.60 & 87.95 \\
\midrule
 \multicolumn{10}{c}{\textit{Train: DeepL + SimAlign - Post-processed Projections}} \\ 
\midrule
    DeepL + SimAlign    & 82.98 & 75.79 & 88.92 & 81.85 & 75.68 &  87.55 & 82.21 & 73.04 & 89.26 \\ 
    DeepL + Awesome     & 82.77 & 75.50 & 88.89 & 81.74 & 75.63 & 87.53 & 82.14 & 73.46 & 89.22 \\
 \midrule
  \multicolumn{10}{c}{\textit{Train: DeepL + Awesome - Post-processed Projections}} \\ 
\midrule
    DeepL + SimAlign     & 83.09 & 75.19 & 89.07 & 81.43 & 75.58 & 87.0 & 82.31 & 73.69 & 89.11 \\ 
    DeepL + Awesome      & 83.04 & 75.29 & 89.05 & 81.32 & 75.53 & 86.98 & 82.23 & 74.09 & 89.07 \\
 \midrule
 \multicolumn{10}{c}{\textit{Train: DeepL + SimAlign - Manually-checked Projections}} \\ 
\midrule
    DeepL + SimAlign    & 83.57 & 75.95 & 90.01 & 80.83 & 75.44 &  86.11 & 82.62 & 74.62 & 88.81 \\
    DeepL + Awesome     & 83.49 & 76.27 & 90.05 & 80.73 & 75.43 & 86.09 & 82.56 & 75.13 & 88.81 \\
 \midrule
 \multicolumn{10}{c}{\textit{Train: DeepL + Awesome - Manually-checked Projections}}  \\ 
\midrule
    DeepL + SimAlign     & 83.49 & 76.84 & 89.09 & 81.22 & 76.20 & 87.37 & 81.96 & 73.32 & 88.54 \\
    DeepL + Awesome      & 83.39 & 77.11 & 89.18 & 81.11 & 76.16 & 87.36 & 81.88 & 73.71 & 88.50 \\
 \bottomrule
\end{tabular}
}
\caption{Data-transfer results with different types of projected data for training. Evaluation tests are manually corrected.}
\label{tab:projection_comparison_data_transfer}
\end{center}
\end{table*}

Comparing the results of training in gold standard English and evaluating on the different types of projected data in Table \ref{table:compare_projections} allowed us to learn that the \emph{post-processed} version is slightly less reliable than testing on the manually-checked projections. Still, differences were minimal. A natural follow-up experiment is to compare the \emph{data-transfer} results when training with the different projection versions and evaluating on the manually-checked data. Results of this experiment are reported in Table \ref{tab:projection_comparison_data_transfer}.\footnote{Appendix \ref{sec:appendix} provides all the detailed results of performing this experiment with different MT systems and word aligners.}

First, it is interesting to observe that while SimAlign performs best when trained on automatic projections, \textsc{awesome} greatly benefits from the post-processing step, allowing it to outperform SigmAlign in both post-processed and manual evaluation settings. This is coherent with the stats reported in Table \ref{table:postprocessed_projection_train_dev} in which post-processing helped to greatly improve \textsc{awesome} automatic projections.

In any case, the most noteworthy aspect to be mentioned is that training on post-processed DeepL + \textsc{awesome} projections obtains practically the same results than when training on manually-checked data. This result is quite interesting as it confirms that \emph{data-transfer} can be fully performed without any manual intervention. As this is the optimal approach in crosslingual transfer for AM, these results suggest that as long as we can obtain quality machine translation and alignments then \emph{data-transfer} should be the choice to perform AM in the medical domain when no manually annotated data is readily available.

\section{Conclusion}\label{sec:conclusion}

In this paper, we investigate crosslingual transfer techniques to perform AM on medical data for a language for which manual annotations are not available. Furthermore, we explore this in a real case scenario in which the only existing dataset annotated for the medical domain was in English. We compare \emph{model-transfer} (apply multilingual MLMs such as mBERT to learn on the available English data and predict in Spanish) with \emph{data-transfer}, which consists of generating training data in Spanish by machine translating and projecting the annotations from English to Spanish.

Contrary to previous results on crosslingual transfer for sequence labelling \cite{garcia2022model}, \emph{data-transfer} clearly is the optimal option for argument component detection. Furthermore, combining the generated Spanish data with the original English set helps to improve results across every evaluation setting, also with respect to the domain-specific models trained with gold-standard English data. Finally, a quick comparison evaluating with the different types of label projections demonstrates that \emph{data-transfer} can be fully performed without manual intervention, a very important result of this study.

Results in relation classification were similar to those of argument component detection except that the multilingual setting does not outperform the results obtained when training on English gold data.

Finally, apart from the scientific findings, we should stress that in this work we have created the first dataset in Spanish to perform Argument Mining in the medical domain. The dataset, fine-tuned models and code will be publicly available upon publication. Additionally, we would like to further explore the method presented in this project to experiment with computational approaches to argumentation for other specific domains and languages for which no annotated data is available.

\section*{Acknowledgments}

This work has been supported by the following MCIN/AEI/10.13039/501100011033 projects: (i) Antidote (PCI2020-120717-2) and EU NextGenerationEU/PRTR (ii) DeepKnowledge (PID2021-127777OB-C21) and by FEDER, EU; (iii) DeepMinor (CNS2023-144375) and European Union NextGenerationEU/PRTR. Anar Yeginbergen's PhD contract is part of the PRE2022-105620 grant, financed by MCIN/AEI/10.13039/501100011033 and by the FSE+.

\bibliographystyle{fullname}

\bibliography{references}

\appendix

\section{Results comparison with different types of projections and MT systems} \label{sec:appendix}

\begin{table*}[h]
\footnotesize{
\begin{center}
\begin{tabular}{c|ccc|ccc|ccc} 
 \toprule
  Test & & Neoplasm &&& Glaucoma &&& Mixed & \\
  \midrule
  & F1 & F1-C & F1-P & F1 & F1-C & F1-P & F1 & F1-C & F1-P \\ 
 \midrule
    DeepL + SimAlign     & 75.87 & 71.21 & 85.70 & 75.64 & 72.08 & 81.18 & 75.25 & 68.84 & 85.08 \\ 
    DeepL + Awesome      & 70.07 & 70.05 & 84.83 & 71.39 & 71.26 & 80.29 & 69.85  & 67.37 & 84.46\\ 
    OPUS-MT + SimAlign      & 77.59 & 71.09 & 86.24 & 75.46 & 69.86 & 80.37 & 	76.63 & 70.26 & 85.50 \\ 
    OPUS-MT + Awesome       & 71.36 & 69.74 & 84.97 & 71.22 & 69.70 & 79.59 & 77.98 & 68.94 & 84.70 \\
 \bottomrule
\end{tabular}
\caption{Model-transfer English to Spanish results of argument components evaluated on automatic projection labels.}
\label{table:arg_comp_zero_auto}
\end{center}
}
\end{table*}


\begin{table*}[h]
\footnotesize{
\begin{center}

\begin{tabular}{c|ccc|ccc|ccc} 
 \toprule
  Test & & Neoplasm &&& Glaucoma &&& Mixed & \\
  \midrule
  & F1 & F1-C & F1-P & F1 & F1-C & F1-P & F1 & F1-C & F1-P \\ 
 \midrule
    DeepL + SimAlign     & 79.99 & 71.69 & 86.67 & 77.30 & 72.29 & 81.51 & 79.25 & 69.36 & 85.85 \\ 
    DeepL + Awesome      & 79.12 & 71.33 & 86.48 & 76.77 & 72.08 & 81.32 & 78.21 & 68.95 & 85.65 \\ 
    OPUS-MT + SimAlign      & 80.68 & 71.54 & 86.69 & 76.81 & 70.08 & 80.59 & 	79.89 & 70.84 & 85.91 \\ 
    OPUS-MT + Awesome       & 80.21 & 71.13 & 86.48 & 76.42 & 69.92 & 80.62 & 79.23 & 70.59 & 85.78 \\
 \midrule
\end{tabular}
\caption{Model-transfer English to Spanish results of argument components evaluated on post-processed projection labels.}
\label{table:arg_comp_post_zero}
\end{center}
}
\end{table*}

\begin{table*}[h]
\begin{center}
\resizebox{15cm}{!}{
\begin{tabular}{c|ccc|ccc|ccc} 
 \toprule
    & & Neoplasm &&& Glaucoma &&& Mixed & \\
  \midrule
Test  & F1 & F1-C & F1-P & F1 & F1-C & F1-P & F1 & F1-C & F1-P \\ 
\midrule
\multicolumn{10}{c}{\textit{Train: OPUS-MT + SimAlign}} \\ 
\midrule
    DeepL + SimAlign     & 82.90 & 75.34 & 87.51 &83.37 & 78.52 &	86.80 & 82.81 & 75.86 & 87.96 \\ 
    DeepL + Awesome      & 82.70 & 75.55 & 87.22 & 83.26 & 78.48 & 86.79 & 82.65 & 75.87 & 87.92 \\ 
    OPUS-MT + SimAlign      & 82.63 & 74.12 & 87.84 & 83.29	& 78.42 & 86.25 & 82.46 & 74.64 & 88.22 \\ 
    OPUS-MT + Awesome       & 82.60 & 74.27 & 87.90  & 83.13 & 78.39 & 86.17 & 82.29 & 74.51 & 88.09 \\ 
 \midrule
\multicolumn{10}{c}{\textit{Train: OPUS-MT + Awesome}} \\ 
\midrule
    DeepL + SimAlign     & 73.32 & 73.67 & 86.87 & 75.36& 73.44 & 85.11 & 74.28 & 73.56 & 87.64 \\ 
    DeepL + Awesome      & 73.15 & 73.74 & 86.73 & 75.32 & 73.40 & 85.11 & 74.26 & 73.99 & 87.63 \\ 
    OPUS-MT + SimAlign   & 74.51 & 73.11 & 87.21 & 75.97 & 73.87 & 84.85 & 74.71 & 73.22 & 87.96 \\ 
    OPUS-MT + Awesome       & 74.41 & 72.91 &87.19 & 75.89 & 73.85 & 84.79 & 74.68 & 73.49 & 87.82 \\
 \midrule
\multicolumn{10}{c}{\textit{Train: DeepL + SimAlign}} \\ 
\midrule
    DeepL + SimAlign    & 82.33 & 75.90 & 87.64 & 80.94 & 75.36 & 85.97 & 81.81 & 74.99 & 88.07 \\ 
    DeepL + Awesome     & 81.98 & 75.27 & 86.99 & 80.84 & 75.31 & 85.96 & 81.73 & 75.36 & 88.03 \\ 
    OPUS-MT + SimAlign  & 81.27 & 74.33 & 87.74 & 80.20 & 74.40 & 85.69 & 80.20 & 71.98 & 87.44 \\ 
    OPUS-MT + Awesome   & 80.99 & 74.15 & 87.71 & 80.04 & 74.37 & 85.61 & 80.11 & 72.24 & 87.30 \\
 \midrule
 \multicolumn{10}{c}{\textit{Train: DeepL + Awesome}} \\ 
\midrule
    DeepL + SimAlign     & 72.78 & 72.66 & 87.03 & 74.21 & 72.82 & 84.74 & 72.40 & 70.17 & 87.99 \\ 
    DeepL + Awesome      & 72.50 & 72.62 & 87.00 & 74.20 & 72.77 & 84.75 & 72.43 & 70.60 & 87.95 \\ 
    OPUS-MT + SimAlign  & 73.74 & 71.53 & 86.85 &  74.18 & 72.89 & 84.14 & 73.25 & 70.49 & 87.58 \\ 
    OPUS-MT + Awesome   & 73.53 & 71.28 & 86.84 &  74.13 & 72.86 & 84.09 & 73.22 & 70.76 & 87.45 \\ 
 \bottomrule
\end{tabular}
}
\caption{Data-transfer results with automatically projected data. Evaluation tests are manually corrected.}
\label{table:app_arg_comp_projected}
\end{center}
\end{table*}

\begin{table*}[h]
\begin{center}
\resizebox{15cm}{!}{
\begin{tabular}{c|ccc|ccc|ccc} 
 \toprule
    & & Neoplasm &&& Glaucoma &&& Mixed & \\
  \midrule
 Test & F1 & F1-C & F1-P & F1 & F1-C & F1-P & F1 & F1-C & F1-P \\ 
\midrule
\multicolumn{10}{c}{\textit{Train: OPUS-MT + SimAlign}} \\ 
\hline
    DeepL + SimAlign     & 83.35 & 74.49 & 87.59 & 82.29 & 75.99 & 86.10 & 83.09 & 74.94 & 87.87 \\ 
    DeepL + Awesome      & 83.07 & 74.81 & 87.02 & 82.19 & 75.95 & 86.09 & 83.02 & 75.31 & 87.83 \\ 
    OPUS-MT + SimAlign      & 83.00 & 73.46 & 88.18 & 81.84 & 76.27 & 85.84 & 82.32 & 72.64 & 88.81 \\ 
    OPUS-MT + Awesome       & 83.08 & 73.99 & 88.34 & 81.72 & 76.25 & 85.77 & 82.19 & 72.68 & 88.68 \\ 
 \midrule
\multicolumn{10}{c}{\textit{Train: OPUS-MT + Awesome}} \\ 
\midrule
    DeepL + SimAlign     & 83.50 & 75.72 & 88.15 & 82.69 & 74.99 & 87.03 & 82.43 & 72.85 & 88.31 \\ 
    DeepL + Awesome      & 83.38 & 75.80 & 88.11 & 82.58 & 74.95 & 87.01 & 82.26 & 72.85 & 88.28 \\ 
    OPUS-MT + SimAlign   & 83.68 & 75.03 & 88.64 & 82.94 & 77.41 & 87.68 & 83.21 & 73.55 & 89.59 \\ 
    OPUS-MT + Awesome       & 83.71 & 75.34 & 88.76 & 82.78 & 77.39 & 87.61 & 83.03 & 73.39 & 89.45 \\   
 \midrule
 \multicolumn{10}{c}{\textit{Train: DeepL + SimAlign}} \\ 
\midrule
    DeepL + SimAlign    & 82.98 & 75.79 & 88.92 & 81.85 & 75.68 &  87.55 & 82.21 & 73.04 & 89.26 \\ 
    DeepL + Awesome     & 82.77 & 75.50 & 88.89 & 81.74 & 75.63 & 87.53 & 82.14 & 73.46 & 89.22 \\ 
    OPUS-MT + SimAlign  & 81.14 & 73.66 & 87.80 & 80.74 & 74.74 & 87.13 & 81.03 & 72.38 & 88.32 \\ 
    OPUS-MT + Awesome   & 81.07 & 73.91 & 87.86 & 80.58 & 74.72 & 87.06 & 80.94 & 72.66 & 88.19 \\
 \midrule
  \multicolumn{10}{c}{\textit{Train: DeepL + Awesome}} \\ 
\midrule
    DeepL + SimAlign     & 83.09 & 75.19 & 89.07 & 81.43 & 75.58 & 87.0 & 82.31 & 73.69 & 89.11 \\ 
    DeepL + Awesome      & 83.04 & 75.29 & 89.05 & 81.32 & 75.53 & 86.98 & 82.23 & 74.09 & 89.07 \\ 
    OPUS-MT + SimAlign  & 81.88 & 73.29 & 88.86 & 80.14 & 74.84 & 85.88 & 81.06 & 73.39 & 88.55 \\ 
    OPUS-MT + Awesome   & 81.74 & 73.04 & 88.86 & 79.99 & 74.82 & 85.82 & 80.97 & 73.65 & 88.42 \\
 \bottomrule
\end{tabular}
}
\caption{Data-transfer results with post-processed projected labels. Evaluation tests are manually corrected.}
\label{table:app_arg_comp_projected_processed}
\end{center}
\end{table*}

\begin{table*}[h]
\begin{center}
\resizebox{15cm}{!}{
\begin{tabular}{c|ccc|ccc|ccc} 
 \toprule
    & & Neoplasm &&& Glaucoma &&& Mixed & \\
  \midrule
  Test & F1 & F1-C & F1-P & F1 & F1-C & F1-P & F1 & F1-C & F1-P \\ 
\midrule
\multicolumn{10}{c}{\textit{Train: OPUS-MT + SimAlign}} \\ 
\midrule
    DeepL + SimAlign     & 83.79 & 75.81 & 88.95 & 81.80 & 74.78 & 86.57 & 82.60 & 74.08 & 88.73 \\ 
    DeepL + Awesome      & 83.50 & 75.90 & 88.61 & 81.69 & 74.73 & 86.55 & 82.53 & 74.46 & 88.69 \\ 
    OPUS-MT + SimAlign      & 83.03 & 74.68 & 88.69 & 82.06 & 75.77 & 87.36 & 82.64 & 72.94 & 89.31 \\ 
    OPUS-MT + Awesome       & 82.80 & 74.58 & 88.69 & 81.90 & 75.73 & 87.28 & 82.51 & 72.94 & 89.18 \\
 \midrule
\multicolumn{10}{c}{\textit{Train: OPUS-MT + Awesome}} \\ 
\midrule
    DeepL + SimAlign     & 83.51 & 75.45 & 88.72 & 82.38 & 76.01 & 87.02 & 82.76 & 74.81 & 88.86 \\
    DeepL + Awesome      & 83.34 & 75.64 & 88.72 & 82.27 & 75.97 & 86.99 & 82.69 & 75.18 & 88.83 \\
    OPUS-MT + SimAlign   & 82.87 & 74.24 & 88.71 & 82.60 & 76.59 & 87.53 & 82.64 & 73.44 & 89.32 \\
    OPUS-MT + Awesome       & 82.66 & 74.07 & 88.69 & 82.44 & 76.55 & 87.45 & 82.55 & 73.70 & 89.18 \\
 \midrule
 \multicolumn{10}{c}{\textit{Train: DeepL + SimAlign}} \\ 
\midrule
    DeepL + SimAlign    & 83.57 & 75.95 & 90.01 & 80.83 & 75.44 &  86.11 & 82.62 & 74.62 & 88.81 \\
    DeepL + Awesome     & 83.49 & 76.27 & 90.05 & 80.73 & 75.43 & 86.09 & 82.56 & 75.13 & 88.81 \\
    OPUS-MT + SimAlign  & 81.82 & 73.98 & 89.49 & 80.14 & 75.70 & 85.75 & 81.06 & 73.71 & 88.67 \\
    OPUS-MT + Awesome   & 81.58 & 73.75 & 89.46 & 80.02 & 75.72 & 85.69 & 80.99 & 74.01 & 88.54 \\
 \midrule
 \multicolumn{10}{c}{\textit{Train: DeepL + Awesome}}  \\ 
\midrule
    DeepL + SimAlign     & 83.49 & 76.84 & 89.09 & 81.22 & 76.20 & 87.37 & 81.96 & 73.32 & 88.54 \\
    DeepL + Awesome      & 83.39 & 77.11 & 89.18 & 81.11 & 76.16 & 87.36 & 81.88 & 73.71 & 88.50 \\
    OPUS-MT + SimAlign  & 82.02 & 75.12 & 88.97 & 80.47 & 75.23 & 86.77 & 81.27 & 73.16 & 88.88 \\
    OPUS-MT + Awesome   & 81.86 & 74.97 & 88.97 & 80.35 & 75.21 & 86.71 & 81.19 & 73.42 & 88.75 \\
 \bottomrule
\end{tabular}
}
\caption{Data-transfer results with manually corrected projections. Evaluation tests are manually corrected.}
\label{table:app_arg_comp_manual_projected}
\end{center}
\end{table*}

\end{document}